# Latent Variable Modeling for Generative Concept Representations and Deep Generative Models

Daniel T. Chang (张遵)

*IBM (Retired)* dtchang43@gmail.com

**Abstract:** Latent representations are the essence of deep generative models and determine their usefulness and power. For latent representations to be useful as generative concept representations, their latent space must support latent space interpolation, attribute vectors and concept vectors, among other things. We investigate and discuss latent variable modeling, including latent variable models, latent representations and latent spaces, particularly hierarchical latent representations and latent space vectors and geometry. Our focus is on that used in variational autoencoders and generative adversarial networks.

## 1 Introduction

*Latent representations* are the essence of *deep generative models (DGMs)* [1 and references therein on DGMs] and determine their usefulness and power. Most studies on DGMs use the Gaussian distribution for both the prior and posterior due to mathematical convenience. It is important to investigate whether the resulting latent representations are appropriate and whether better latent representations would result by using a different distribution or a different latent variable model.

*Generative concept representations* [1-2] are latent representations learned using DGMs. A key characteristic of generative concept representations is that they can be directly manipulated to generate new concepts with desired attributes. Therefore, for latent representations to be useful as generative concept representations, their *latent space* must support *latent space interpolation*, attribute vectors and concept vectors, among other things. *Attribute vectors* enable latent space manipulation at the attribute level, whereas *concept vectors* enable *latent space manipulation* at the concept level.

The nature and characteristics of latent representations, and their latent space, learned in a DGM are determined by the *latent variable model* used in the DGM, the sample data representations, and the DGM architecture and algorithms. In this paper, we investigate and discuss latent variable modeling, including latent variable models, latent representations and latent spaces, particularly hierarchical latent representations and latent space vectors and geometry. Our focus is mainly on that used in *prescribed DGMs* [1, 3], particularly *variational autoencoders (VAEs)* [1 and references therein on VAEs]. *Implicit DGMs* [1, 3], e.g., *generative adversarial networks (GANs)* [1 and references therein on GANs], are discussed separately in the last section.

# 2 Latent Variable Models

From the perspective of generative concept representations, a *latent variable model (LVM)* is a *probabilistic graphical model (PGM)* [1 and references therein on PGMs] of *observed variables x* that incorporates *latent variables z*. The inclusion of latent variables allows us to capture hidden dependencies among observed variables and thus learn the structure underlying the data generating mechanism. More importantly, the latent variables can provide an alternative, low-dimensional (due to dimension reduction) representation for the observed variables, thus accomplishing *representation learning*.

For *prescribed DGMs*, an LVM is prescribed by the *joint distribution $p_\theta(x, z)$* with parameter $\theta$. The model can be constructed in two steps [4]. In the first step, we factorize the joint distribution into a product of *conditional and marginal distributions*. Through conditioning, we introduce dependencies that transform the joint distribution into a mixture of distributions. Implicit in this mixture is a data-generating process that gives rise to the sample data. In the second step, we specify *particular distributions* to the factor distributions.

## 2.1 LVM Decomposition

There are a large number of ways we can decompose the joint distribution into a product of factor distributions [4]. Each decomposition implicitly encodes a particular number of modeling assumptions.

*Shallow LVMs*

A *shallow LVM,* or *flat LVM,* has a single latent layer. With a single layer, the joint distribution is decomposed for the *generative model* as [3, 5]:

$$p_\theta(x, z) = p_\theta(z)\, p_\theta(x \mid z),$$

and for the *inference model* as:

$$p_\theta(x, z) = p_\theta(x)\, p_\theta(z \mid x).$$

*Learning* from data occurs through the transformation of the *prior $p_\theta(z)$* (defined before observing the data) into the posterior $p_\theta(z \mid x)$ (obtained after observing the data) while maximizing the marginal log-likelihood $\log(p_\theta(x))$. The *generative process* is $z \sim p_\theta(z)$; $x \sim p_\theta(x \mid z)$.

In contrast to shallow LVMs, *deep LVMs* have multiple latent layers.

## Hierarchical LVMs

A *hierarchical LVM* is a deep LVM, whose joint distribution is decomposed for the *generative model* in a *hierarchical, top-down* manner [3, 6]:

$$p_\theta(\mathbf{x}, \mathbf{z}) = p_\theta(\mathbf{z}_L) \left(\prod_{i=L-1}^{1} p_\theta(\mathbf{z}_i | \mathbf{z}_{i+1})\right) p_\theta(\mathbf{x} | \mathbf{z}_1),$$

where L is the number of latent layers and $\mathbf{z}_i$ (i = 1 … L) are the latent variables at the i-th layer.

The inference model and learning for hierarchical LVMs vary. See 3.3 Hierarchical Latent Representations for discussion.

## Deep Exponential-Family LVMs

*Deep exponential-family LVMs (DEF LVMs)* [7] are a class of LVMs that are inspired by the hidden structures used in *deep neural networks*. DEF LVMs capture *a hierarchy of dependencies between latent variables*, and are easily generalized to many settings through *exponential families* (see 2.2 Latent Variable Distributions).

To construct DEF LVMs, we *chain exponential families together in a hierarchy*, where the draw from one layer controls the natural parameters of the next. For each data point $\mathbf{x}_n$, the model has *L layers of hidden variables {$z_{n,1}$, ..., $z_{n,L}$}* where each $\mathbf{z}_{n,l} = \{z_{n,l,1}, ..., z_{n,l,K(l)}\}$. Shared across data, the model has *L - 1 layers of weights {$W_1$, ..., $W_{L-1}$}*, where each $\mathbf{W}_l = \{w_{l,1}, ..., w_{l,K(l)}\}$. We assume the weights have a prior distribution $p(\mathbf{W}_l)$.

The *generative model* uses a top-down process and the distribution of a single data point $\mathbf{x}$ (with n omitted for simplicity) is obtained as follows. First, the top layer of latent variables are drawn given a natural parameter $\boldsymbol{\theta}$:

$$p(z_{L,k}) = \mathbf{EF}_L(z_{L,k}, \boldsymbol{\theta}),$$

where $\mathbf{EF}$ denotes an exponential family. Next, each latent variable is drawn conditional on the previous layer:

$$p(z_{l,k} | \mathbf{z}_{l+1}, \mathbf{w}_{l,k}) = \mathbf{EF}_l(z_{l,k}, g_l(<\mathbf{z}_{l+1}, \mathbf{w}_{l,k}>)),$$

where $g_l$, the *link function*, maps the inner product to the natural parameter **θ**. Note *each of the K(l) variables in layer l depends on all the variables of the higher layer* (which gives the flavor of a deep neural network) and *the type of exponential family can change across layers*. The data are drawn conditioned on the lowest layer, $p(\mathbf{x} \mid \mathbf{z}_1)$.

The inference model and learning for DEF LVMs are discussed in 3.3 Hierarchical Latent Representations.

## 2.2 Latent Variable Distributions

*Gaussian Distribution*

The multivariate *Gaussian (or Normal) distribution* is the most common distribution used in VAEs. It is parameterized by the *mean **μ*** and the *covariance **Σ*** (or with simplification the *diagonal covariance $\sigma^2 \mathbf{I}$*).

The Gaussian distribution has been used in both shallow LVMs and hierarchical LVMs. For shallow LVMs, it is used for the prior and posterior [3, 5]. For hierarchical LVMs, it is used for the prior and posterior at each layer [3, 6]. In either case, the Gaussian distribution can be used for the likelihood for real-valued observed variables.

*von Mises-Fisher Distribution*

Some data types like *directional data* (e.g., wind direction) are better modeled through spherical representations. The *von Mises-Fisher (vMF) distribution* [8] is often seen as the Gaussian distribution on a *hypersphere*. Analogous to the Gaussian distribution, the vMF distribution is parameterized by the *mean direction μ* and the *concentration κ* around μ.

The vMF distribution has been used in shallow LVMs. It is used for both the prior and posterior [8-9]. With the choice of κ as a fixed model hyper-parameter for the posterior [9], the KL divergence term only depends on the variance of the vMF distribution. Doing so not only averts the *KL collapse* [10], but also gives better likelihoods than the Gaussian distribution across a range of modeling conditions.

*Exponential Families*

*Exponential families* [7, 11] are an important class of distributions which encompass most common statistical distributions and are the only ones that have *sufficient statistics* and allow us to perform data reduction for inference problem. An exponential family is a set of probability distributions admitting the following canonical decomposition:

$$p(\mathbf{x}; \boldsymbol{\theta}) = \exp(<t(\mathbf{x}), \boldsymbol{\theta}> - F(\boldsymbol{\theta}) + k(\mathbf{x}))$$

where

- $t(\mathbf{x})$ is the sufficient statistic,

- $\boldsymbol{\theta}$ are the natural parameters,

- $<\ldots>$ is the inner product,

- $F(\cdot)$ is the log-normalizer,

- $k(\mathbf{x})$ is the carrier measure.

The order D of the family is the dimension of the natural parameter space.

Both the *Gaussian distribution* and the *vMF distribution* are members of exponential families. Exponential families, as a whole, are the building blocks of DEF LVMs, discussed previously.

## 3 Latent Representations

The nature and characteristics of latent representations learned in a DGM are determined by the LVM used in the DGM, the sample data representations, and the DGM architecture and algorithms.

### 3.1 Sample Data Representations

The sample data representation determines the content of the latent representation learned from it. Different data representations of the same sample data can therefore affect the quality and usefulness of the latent representations learned.

A good example is provided by the latent representations learned for molecular structures coded in SMILES, as discussed in [1]. The same SMILES sample data can be represented as strings, parse trees or attributed parse trees and use CVAE, GVAE or SD-VAE, respectively, to learn the latent representations. The resulting latent representations have increased syntactic and semantic constraints, respectively, and thus generate outputs with increased syntactic and semantic validity.

## 3.2 Disentangled Latent Representations

Standard VAEs typically produce entangled latent representations, which encode all features of the data into a single latent variable, and the latent representations usually do not exhibit consistent meaning along axes of variation. *Disentangled latent representations* [12-14] encode distinct aspects of the data into *separate latent variables*, providing consistent meaning along *axes of variation*. They offer several *benefits* such as ease of deriving invariant representations to nuisance factors, transferability to other tasks, interpretability, etc.

*Semi-supervised Disentanglement*

In *semi-supervised disentanglement* [12] a general class of *partially-specified probabilistic graphical model*s is used, in which we only need specify the exact relationship for some *subset of the latent variables* in the model. This allows us to specify precisely those *axes of variations (and their dependencies)* we have information about or would like to extract, and learn *disentangled representations* for them, while leaving the rest to be learned in an entangled manner. A subclass of partially-specified models that is particularly common is that where we can obtain *supervision data* for some subset of the variables.

Both the generative model $p_\theta(x, y, z)$ and the approximate posterior $q_\varphi(y, z \mid x)$ can have arbitrary conditional dependency structures involving latent variables defined over a number of different distribution types. The subset of latent variables *y are disentangled* with specified conditional dependency structures, whereas the subset *z are entangled*. For disentangled variables, we assume that *supervision labels* are available for some fraction of the data, e.g., a dataset with N unsupervised data points $D = \{x_1, …, x_N\}$ and M supervised data points $D_{sup} = \{(x_1, y_1), …, (x_M, y_M)\}$.

A generalized formulation of *semi-supervised learning with VAEs* is used that enables the framework to automatically employ the correct factorization of the *variational inference objective* for any given choice of model and set of latent variables taken to be disentangled. The objective consists of a *supervised term* in addition to the standard *unsupervised term*, with a *constant γ* that controls the *relative strength* of the supervised term.

*Grouped Latent Representations*

*Group latent representations* allow us to capture *grouping semantics* in the data where within a group the samples share a common factor of variation. For example, consider a data set of objects (which are concept instances with the object class being the concept) with two factors of variation (each factor being an attribute): shape and color. A possible grouping organizes the objects by shape; another possible grouping organizes the objects by color.

The *Multi-Level VAE (ML-VAE)* [13] learns a disentangled representation of *grouped data*. It separates the latent representation into *semantically relevant parts* by working both at the group level and the observation level. *Group-level supervision* is used at training to organize observations into *groups*, where within a group the observations share a common but unknown value for one of the *factors of variation*. By anchoring the semantics of the grouping into a disentangled representation, the ML-VAE enables *manipulation* of the latent representation *based on groups*. For example, for the above data set of objects, we can manipulate the latent representation to perform operations such as swapping the shape to generate new objects.

*Disentangled Inferred Prior VAE*

The majority of approaches to learning disentangled latent representations, such as the previous ones, assume some sort of supervision. However, the major issue with using supervision is that, in most real-world scenarios, we only have access to raw observations without any supervision about the factors of variation.

The *Disentangled Inferred Prior VAE (DIP-VAE)* [14] is a principled variational framework for inferring disentangled latent variables from *unlabeled observations*. Disentanglement is encouraged by introducing a *regularizer over the inferred prior*.

We define the *inferred prior* or *expected variational posterior* as:

$$q_\varphi(z) = \int q_\varphi(z \mid x) p_\theta(x) dx$$

Minimizing $D_{KL}(q_\varphi(z) \parallel p_\theta(x))$ or any other suitable distance $D(q_\varphi(z), p_\theta(x))$ explicitly will give us better control on the disentanglement. This motivates us to add $D(q_\varphi(z), p_\theta(x))$ as part of the objective to encourage disentanglement during inference:

$$\sum_z q_\varphi(z \mid x) \log(p_\theta(x \mid z)) - D_{KL}(q_\varphi(z \mid x) \parallel p_\theta(z)) - \lambda D(q_\varphi(z), p_\theta(x))$$

where λ controls its contribution to the overall objective.

## 3.3 Hierarchical Latent Representations

*Hierarchical latent representations* allow us to capture *complex structure* in the data. This is important since a latent representation that has some factorized structure, and consistent semantics associated to different parts, is more likely to be of general use. In other words, hierarchical latent representations are *hierarchically disentangled* and offer the benefits of disentangled latent representations.

*Hierarchical LVMs* are highly expressive and flexible. They would seem to be a natural choice to use for learning hierarchical latent representations. However, hierarchical LVMs are *difficult to optimize* for deep hierarchies due to multiple layers of conditional stochastic layers, as discussed below.

Therefore, it is an *open question* as to *what type of LVM* (shallow, hierarchical, deep exponential families, …)*, what kind of latent variable distribution* (simple unimodal, complex multimodal, …) and *what sort of inference model* (bottom up, top down, …) are best to use for learning hierarchical latent representations.

### Hierarchical VAE

The *Hierarchical VAE (HVAE)* [6, 15-16] uses the *hierarchical LVM,* with the Gaussian distribution, for learning hierarchical latent representations. Whereas the *generative model* of hierarchical LVMs is parameterized as a *top-down process*, as discussed previously, the HVAE *inference model* is parameterized as a *bottom-up process*:

$$q_\varphi(\mathbf{z} \mid \mathbf{x}) = q_\varphi(\mathbf{z}_1 \mid \mathbf{x}) \left(\prod_{i=2}^{L} q_\varphi(\mathbf{z}_i \mid \mathbf{z}_{i-1})\right)$$

The inference and generative distributions are *computed separately* with no explicit sharing of information.

The HVAE has two limitations [15-16]. First, in theory, if the hierarchical LVM can be trained to optimality, then the bottom layer alone contains enough information to reconstruct the data distribution, and the layers above the first one can be ignored. In practice, *higher layers are not utilized*. Second, many of the building blocks (e.g., the Gaussian distribution) commonly used in the HVAE are unlikely to learn *disentangled features*.

### Ladder VAE

The *Ladder VAE (LVAE)* [15] uses the *hierarchical LVM,* with the Gaussian distribution, for learning hierarchical latent representations. LVAE and HVAE only differ in the *inference model*, though they have similar number of parameters, whereas the *generative models* are identical.

The *LVAE inference model* recursively corrects the generative distribution with a data dependent approximate likelihood term. First a *deterministic upward pass* computes the approximate likelihood. This is followed by a *stochastic downward pass* recursively computing both the approximate posterior and generative prior distributions:

$$q_\varphi(z \mid x) = q_\varphi(z_L \mid x) \left( \prod_{i=L-1}^{1} q_\varphi(z_i \mid z_{i+1}) \right)$$

The same top-down dependency structure is used both in the inference and generative model.

The approximate posterior distribution can be viewed as merging information from a bottom up computed approximate likelihood with top-down prior information from the generative distribution. The sharing of information (and parameters) with the generative model gives the inference model knowledge of the current state of the generative model in each layer and the top down-pass recursively corrects the generative distribution with the data dependent approximate log-likelihood.

The learned *latent representations* differ qualitatively between the LVAE and HVAE with the LVAE learning both a *deeper and more distributed* representation.

*Variational Ladder Autoencoder*

The *Variational Ladder Autoencoder (VLAE)* [16] uses the *shallow LVM,* with the Gaussian distribution, for learning hierarchical latent representations. It is able to learn highly *interpretable and disentangled hierarchical features* by crafting a *network architecture* that prefers to place high-level features on certain parts of the latent variables, and low-level features on others.

The basic assumption is: if $z_i$ is more abstract than $z_j$, then the inference mapping $q(z_i \mid x)$ and generative mapping when other layers are fixed $p(x \mid z_i, z_{\neg i} = z^0_{\neg i})$ require a more expressive network to capture. This assumption suggests that we should *use neural networks of different depth* to learn the corresponding features / latent variables; the more abstract features require deeper networks, and vice versa.

The VLAE uses essentially the inference and learning framework for a shallow-LVM VAE; *the hierarchy is only implicitly defined by the network architecture* used for learning shallow (flat) latent variables.

*Deep Exponential Families*

*Deep exponential families (DEFs)* [7] use the DEF LVM for learning hierarchical latent representations. DEFs combine the powerful representations of *deep neural networks* with the flexibility of *probabilistic graphical models*.

DEFs use variational inference which seeks to minimize the KL divergence to the posterior p from an approximate distribution q. This is equivalent to maximizing the following:

$$L(q) = E_{q(\mathbf{z},\mathbf{W})}[\log p(\mathbf{x}, \mathbf{z}, \mathbf{W}) - \log q(\mathbf{z}, \mathbf{W})]$$

where E is the expectation. Let N be the number of observations, then

$$q(\mathbf{z}, \mathbf{W}) = q(\mathbf{W}_0) \prod_{l=1}^{L} q(\mathbf{W}_l) \prod_{n=1}^{N} q(\mathbf{z}_{n,l})$$

Each component in $q(\mathbf{z}_{n,l})$ is

$$q(z_{n,l,k}) = \mathbf{EF}_l(z_{n,l,k}, \lambda_{n,l,k})$$

where $\lambda$ are the natural parameters and the exponential family $\mathbf{EF}_l(.)$ is the same one as that in the generative model distribution p. Similarly, we choose $q(\mathbf{W}_l)$ to be in the same family as $p(\mathbf{W}_l)$ with parameters $\xi$.

DEFs have been applied using 28 different DEFs instances varying the number of layers (1, 2 or 3), the latent variable distributions (gamma, Poisson, Bernoulli) and the weight distributions (Gaussian, gamma) using a Poisson observational model. The results show improvements over strong baselines for both *topic modeling* and *collaborative filtering* on a total of four corpora.

## 4 Latent Spaces

The nature and characteristics of the latent space produced in a DGM are determined mainly by the LVM used in the DGM and the DGM architecture and algorithms.

## 4.1 Latent Space Interpretation

*Latent space interpretation* is critical to generic concept representations for use in *creative applications*. For creative purposes, we desire latent space interpolations that are *smoothly varying* and *semantically meaningful* [17]. First, given a point in latent space which maps to a sample data point, points near it in latent space should map to data points which are semantically similar. Second, we additionally desire that the latent space disentangles meaningful semantic groups in the dataset.

If $z_1$ and $z_2$ are the latent points corresponding to data points $x_1$ and $x_2$, then we can perform latent space interpolation by:

$$z_\alpha = f(z_1, z_2; \alpha)$$

for $\alpha \in [0, 1]$, where f is the *interpolation function*. Our goal for the interpolation to be smoothly varying and semantically meaningful is satisfied if (1) $p_\theta(x \mid z_\alpha)$ is a realistic data point for all $\alpha$, (2) $p_\theta(x \mid z_\alpha)$ transitions in a semantically meaningful way from $p_\theta(x \mid z_1)$ to $p_\theta(x \mid z_2)$ as we vary $\alpha$ from 0 to 1, and (3) $p_\theta(x \mid z_\alpha)$ is perceptually similar to $p_\theta(x \mid z_{\alpha+\delta})$ for small $\delta$.

*Linear Interpolation*

Linear interpolation is frequently used since it is easily understood and implemented. We can perform *linear interpolation* in latent space by computing [17]

$$z_\alpha = (1-\alpha)z_1 + \alpha z_2$$

for $\alpha \in [0, 1]$. This treats the interpolation as *a line path in an n-dimensional Euclidean space*.

*Spherical Linear Interpolation*

When the prior of a VAE is a Gaussian distribution, in high dimensional spaces (> 50 dimensions) samples from the prior are practically indistinguishable from samples from the uniform distribution on the unit hypersphere. In such a space, linear interpolation traverses locations that are extremely unlikely given the prior. In practice we therefore use *spherical linear interpolation (slerp)* [18]:

$$z_\alpha = \mathrm{Slerp}(z_1, z_2; \alpha) = \frac{\sin(1-\alpha)\theta}{\sin\theta} z_1 + \frac{\sin\alpha\theta}{\sin\theta} z_2$$

for α ϵ [0, 1]. This treats the interpolation as *a circle path on an n-dimensional hypersphere*.

*Geodesic Interpolation*

This is discussed in 4.3 Latent Space Geometry.

## 4.2 Latent Space Vectors

*Latent space vectors* are critical to generic concept representations for use in *creative applications*. They consist of attribute vectors and concept vectors. *Attribute vectors* enable latent space manipulation at the *attribute level*, whereas *concept vectors* enable latent space manipulation at the *concept level*. Latent space vectors allows *high-level abstract features* (semantic attributes and semantic concepts), which represent many separate low-level features simultaneously, to be modified or created using simple vector arithmetic.

*Attribute Vectors*

*Attribute vectors* [17, 19-20] are based on the idea that certain directions in the latent space should correspond to particular semantic *attributes* (e.g., mustache), which are characteristics of semantic *concepts* (e.g., person), along which the sample data distribution varies. This assumes, of course, that the latent space is properly *disentangled*, as discussed previously or using the *VAE/GAN* [19].

A simple method for calculating attribute vector is [19]: For each attribute, we compute the *mean vector* for samples with the attribute and the mean vector for samples without the attribute. We then compute the attribute vector as the *difference* between the two mean vectors. This method works for attributes that are not highly correlated with others.

The balancing technique [20] can be applied to attributes that are highly correlated. It uses replication on the training data such that the dataset is balanced across attributes. Decoupling attributes allows individual effects to be applied separately.

*Concept Vectors*

*Concept vectors* can be computed as the *mean vector* for concept samples [17], if available. Otherwise, they can be computed as the *sum vector* of corresponding attribute vectors, with the assumption that concepts are the aggregate of their attributes.

Concept vectors in the form of *object vectors* [21] have been used successfully for 3D object generation and learning from 2D object images. Three methods are used for understanding the latent space of object vectors. First, *visualize* what an individual dimension of the object vector represents; second, explore the possibility of *interpolating between two object vectors* and observe how the generated objects change; last, apply *shape arithmetic* in the latent space.

*Vector Arithmetic*

*Vector arithmetic* [17, 19-20] can be applied to attribute vectors or concept vectors to produce samples that reflect changes in these attributes or concepts, respectively. The following are example vector arithmetic which has been applied.

*Analogy* [20] has been shown to capture regularities in continuous latent spaces. Analogies are usually written in the form:

A : B :: C : ?

which asks: "What is the result of applying the transformation A : B to C?" This can be solved using the vector math:

? = C + B – A

*Attribute swapping* [17] involves starting with a *concept vector*, subtracting an *attribute vector* corresponding to the concept vector, and adding another attribute vector. Decoding the resulting concept vector can produce a realistic realization of the initial concept sample with the attribute swapped. This can also be used to *test the latent space's ability to produce reliable concept vectors and attribute vectors* [17] and, therefore, whether the latent space is useful for generative concept representations.

### 4.3 Latent Space Geometry

DGMs provide a *nonlinear generator (decoder) function* that maps latent points into the data space and, in the case of VAEs, a *nonlinear inference (encoder) function* that maps data points into the latent space. The nonlinearity of the function(s) implies that *the latent space gives a distorted view of the data space*. Therefore, whereas the data space is generally treated as a flat Euclidean space, the latent space should not be seen as a flat Euclidean space, but rather as a *curved Riemannian space* [23-25]. This not only improves our understanding of the latent space, but also improves latent space interpolations, distance metrics and clustering, latent probability distributions, sampling algorithms, interpretability, and generation of new but valid data [24].

*Riemannian Geometry*

The basis of *Riemannian geometry* [22] is Riemann's great insight that *line elements* could be used as the starting point for the consideration of a geometry. An n-dimensional *Riemannian space* is a space in which the line element takes the general form

$$dl^2 = \sum_{i,j=1}^{N} g_{ij} \, dx^i \, dx^j,$$

where $dx^1, dx^2, \ldots, dx^n$ are the differentials of the n coordinates that describe the space, and the various $g_{ij}$ are functions of the coordinates known as *metric coefficients* that are required to be symmetric in the sense that $g_{ij} = g_{ji}$.

The complete set of metric coefficients $[g_{\mu\nu}]$ is called the *metric*, or the *metric tensor*, G. Once you know the metric, the geometry of the space is entirely determined. However, the geometry does not uniquely determine the metric because there are many possible coordinate systems and hence many different ways of writing the metric.

*Parallel transport* and *connection coefficients* are important in several contexts, particularly in connection with *differentiation* in curved spaces. They also provide an important indicator of the *curvature* of a space. Given the components $v^i$ of a vector at some point on a curve specified by $x^i(u)$ in a Riemannian space with coordinates $x^1, \ldots, x^n$ and metric $[g_{ij}]$, the components of the *parallel transported vector* at some neighboring point on the curve are given by:

$$v^i(u + du) = v^i(u) - \sum_{j,k} \Gamma^i_{jk} \, v^j \frac{dx^k}{du} \, du$$

where the *connection coefficient* $\Gamma^i_{jk}$ is given by:

$$\Gamma^i_{jk} = \frac{1}{2} \sum_l g^{il} \left( \frac{\partial g_{lk}}{\partial x^j} + \frac{\partial g_{jl}}{\partial x^k} - \frac{\partial g_{jk}}{\partial x^l} \right)$$

The analogues of straight lines (flat space) and minor arcs of great circles (hypersphere) in a Riemannian space are referred to as *geodesics*. A geodesic satisfies the geodesic equations:

$$\frac{d^2x^i}{d\lambda^2} + \sum_{j,k} \Gamma^i{}_{jk} \frac{dx^j}{d\lambda} \frac{dx^k}{d\lambda} = 0$$

where λ is the affine parameter.

The *curvature* of a Riemannian space is defined by the *Riemann curvature tensor* or *Riemann tensor*:

$$R^l{}_{ijk} \equiv \frac{\partial \Gamma^l{}_{ik}}{\partial x^j} - \frac{\partial \Gamma^l{}_{ij}}{\partial x^k} + \sum_m \Gamma^m{}_{ik}\Gamma^l{}_{mj} - \sum_m \Gamma^m{}_{ij}\Gamma^l{}_{mk}$$

The Riemann tensor has many symmetry involving its indices. These symmetries reduce the number of independent components to 20 in four-dimensional spaces and only 6 in three-dimensional spaces.

*Riemannian Manifold*

The latent space learned using DGMs is a curved Riemannian space, or more precisely a Riemannian manifold. A *Riemannian manifold* is a differentiable manifold M with a metric tensor G. (A *differential manifold* is a space that can be given coordinates locally; the relations between any two coordinate systems are differentiable.)

A DGM represents a *mapping, g : Z -> X*, from some low-dimensional latent space $Z \subseteq R^d$ to a high-dimensional data space $X \subseteq R^D$ (typically, d << D). g is a smooth manifold, $M \subset X$. For every point $z \in Z$, the *Jacobian* of g at **z**, $J_g(\mathbf{z})$, has rank d and components: $J_g^{i,j} = \frac{\partial g_i}{\partial z_j}$. The *Riemannian metric* of the manifold then is given by the formula [23]:

$$G(\mathbf{z}) = J_g(\mathbf{z})^T J_g(\mathbf{z}).$$

*Geodesics*

The geodesic equations are a system of ordinary differential equations. Their solution requires numerical integration and involves expensive computations, which cannot be readily accommodated using standard frameworks for deep learning. *Efficient algorithms for computing geodesics* have been proposed in [23, 26].

*Geodesic interpolation* [23] can be carried out in two steps: (1) compute the geodesic curve connecting a given pair of latent points on the manifold M and (2) discretize the geodesic curve at 1/α points for α ∈ (0, 1].

Given a geodesic path from a point A ϵ M to a point B ϵ M, we can transfer the change from A -> B into a change of a third point C ϵ M. This type of *geodesic analogy* [23] is similar to the analogy in Euclidean space discussed previously, and can be performed in three steps: (1) compute the initial velocity to the geodesic from A to B, (2) parallel transport this velocity along the geodesic from A to C, and (3) use this velocity at C to compute a geodesic step.

# 5 Implicit DGMs

For *implied DGMs*, the generative model is defined by the *generator function* $x = g(z; \theta^{(g)})$, which assumes an implicit joint distribution $p_\theta(x, z)$. If the generator function is invertible, $p_\theta(x, z)$ has a closed form, same as prescribed DGMs. Otherwise, $p_\theta(x, z)$ is intractable and how best to perform the *inverse mapping from data space to latent space* remains an *open research problem*. The following discuss some recent work on the inverse mapping for GANs.

## 5.1 Extending the GAN Architecture for the Inverse Mapping

In their standard form, GANs provide no means of learning the inverse mapping. The following approaches provide a means of learning the inverse mapping by extending the GAN architecture.

In the *Bidirectional Generative Adversarial Networks (BiGAN)* [27], in addition to the *generator g(z)* from the standard GAN, it includes an *encoder e(x)* which maps data **x** to latent representations **z**. The BiGAN *discriminator d(x, z)* discriminates not only in data space (**x** versus g(**z**)), but *jointly in data and latent space* (tuples [**x**, e(**x**)] versus [g(**z**), **z**]), where the latent component is either an encoder output e(**x**) or a generator input **z**. The generator and encoder must learn to invert one another in order to fool the BiGAN discriminator.

The *Adversarially Learned Inference (ALI)* model [28] uses the same approach as BiGAN. It jointly learns a generation network and an inference network using an adversarial process. The *generation network (decoder)* generates data samples from the latent space to the data space while the *inference network (encoder)* maps training samples in the data space to the latent space. An adversarial game is cast between these two networks and a *discriminative network* is trained to distinguish between *joint latent/data-space samples* from the generative network and joint samples from the inference network. ALI's objective is to match the decoder joint distribution p(**x**, **z**) with the encoder joint distribution q(**x**, **z**). If this is achieved, then we can be ensured that the *conditional q(z | x)* matches the posterior p(**z** | **x**), thus achieving the inverse mapping.

## 5.2 Other Approaches to the Inverse Mapping for GANs

Typically, given a target sample generated by a GAN, inversion is achieved by iteratively finding a latent point **z** which, when passed through the GAN generator, produces a sample that is very similar to the target sample. A simple gradient-based technique called *stochastic clipping* [29] can be used to achieve precise reconstruction of latent points **z** for deep convolutional GANs with image data.

A direct inversion approach, however, is used by the algorithm *inversion* [30], which projects target samples, specifically images, to the latent space using a pre-trained GAN, provided that the computational graph for the GAN is available. The inferred latent point **z,** when passed through the GAN generator, produces a sample visually similar to the target sample.

## 6 Summary and Conclusion

Generative concept representations are latent representations learned using deep generative models. A key characteristic of generative concept representations is that they can be directly manipulated to generate new concepts with desired attributes. Therefore, for latent representations to be useful as generative concept representations, their latent space must support latent space interpolation, attribute vectors and concept vectors, among other things. As such, we investigated and discussed latent variable modeling, including latent variable models, latent representations and latent spaces.

Hierarchical latent representations allow us to capture complex structure in the data. They are hierarchically disentangled and offer the benefits of disentangled latent representations. Unfortunately, it is an open question as to what type of latent variable model, what kind of latent variable distribution and what sort of inference model are best to use for learning hierarchical latent representations.

Attribute vectors and concept vectors are critical to generic concept representations for use in creative applications. They allow high-level abstract features (semantic attributes and semantic concepts), which represent many separate low-level features simultaneously, to be modified or created using simple vector arithmetic. If the latent space can produce reliable attribute vectors and concept vectors, we have taken a big step toward generative concept-oriented deep learning.

Due to the nonlinearity of deep generative models, it is critical to recognize that the latent space should not be seen as a flat Euclidean space, but rather as a curved Riemannian manifold. This not only improves our understanding of the latent space, but also improves latent space interpolations, distance metrics and clustering, latent probability distributions, sampling algorithms, interpretability, and generation of new but valid data.

# References

[1] Daniel T. Chang, "Concept-Oriented Deep Learning: Generative Concept Representations," arXiv preprint arXiv:1811.06622 (2018).
[2] Daniel T. Chang, "Concept-Oriented Deep Learning," arXiv preprint arXiv:1806.01756 (2018).
[3] Zhijian Ou, "A Review of Learning with Deep Generative Models from Perspective of Graphical Modeling," arXiv preprint arXiv:1808:01630 (2018).
[4] Rick Farouni, "A Contemporary Overview of Probabilistic Latent Variable Models," arXiv preprint arXiv:1706.08137 (2017).
[5] Diederik P Kingma and Max Welling, "Auto-encoding Variational Bayes," ICLR, 2014.
[6] Danilo Jimenez Rezende, Shakir Mohamed, and DaanWierstra, "Stochastic Backpropagation and Approximate Inference in Deep Generative Models," ICML, 2014.
[7] Rajesh Ranganath, Linpeng Tang, Laurent Charlin, and David Blei, "Deep Exponential Families," in Artificial Intelligence and Statistics, pages 762–771, 2015.
[8] T. R. Davidson, L. Falorsi, N. D. Cao, T. Kipf, and J. M. Tomczak, "Hyperspherical Variational Autoencoders," arXiv preprint arXiv:1804.00891 (2018).
[9] J. Xu and G. Durrett, "Spherical Latent Spaces for Stable Variational Autoencoders," in Empirical Methods in Natural Language Processing (EMNLP), 2018.
[10] S. R. Bowman, L. Vilnis, O. Vinyals, A.M. Dai, R. Jozefowicz, and S. Bengio, "Generating Sentences from a Continuous Space," arXiv preprint arXiv:1511.06349 (2015).
[11] F. Nielsen and V. Garcia, "Statistical Exponential Families: A Digest with Flash Cards," arXiv preprint arXiv:0911.4863 (2011).
[12] N. Siddharth, B. Paige, J.-W. van de Meent, A. Desmaison, N. D. Goodman, P. Kohli, F. Wood, and P. H. S. Torr, "Learning Disentangled Representations with Semi-Supervised Deep Generative Models," NIPS, 2017.
[13] D. Bouchacourt, R. Tomioka, and S. Nowozin, "Multi-level Variational Autoencoder: Learning Disentangled Representations from Grouped Observations," arXiv preprint arXiv:1705.08841 (2017).
[14] A. Kumar, P. Sattigeri, and A. Balakrishnan, "Variational Inference of Disentangled Latent Concepts from Unlabeled Observations," in ICLR, 2018.
[15] C. K. Sønderby, T. Raiko, L. Maaløe, S. K. Sønderby, and O. Winther, "Ladder Variational Autoencoders," arXiv preprint arXiv:1602.02282 (2016).
[16] S. Zhao, J. Song, S. Ermon, "Learning Hierarchical Features from Generative Models," arXiv preprint arXiv:1702.08396 (2017).
[17] Adam Roberts, Jesse Engel, Colin Raffel, Curtis Hawthorne, and Douglas Eck, "A Hierarchical Latent Vector Model for Learning Long-term Structure in Music," arXiv preprint arXiv:1803.05428 (2018).
[18] Tom White, "Sampling Generative Networks," arXiv preprint arXiv:1609.04468 (2016).
[19] A. B. L. Larsen, S. K. Sønderby, and O. Winther, "Autoencoding Beyond Pixels Using A Learned Similarity Metric," arXiv preprint arXiv:1512.09300 (2015).
[20] T. White, "Sampling Generative Networks," arXiv preprint arXiv:1609.04468 (2016).
[21] J. Wu, C. Zhang, T. Xue, B. Freeman, and J. Tenenbaum, "Learning A Probabilistic Latent Space of Object Shapes via 3d Generative-Adversarial Modeling," in NIPS, pages 82–90, 2016.
[22] Robert J. A. Lambourne, *Relativity, Gravitation and Cosmology* (Cambridge University Press, 2010).
[23] Hang Shao, Abhishek Kumar, and P. Thomas Fletcher, "The Riemannian Geometry of Deep Generative Models," arXiv preprint arXiv:1711.08014 (2017).
[24] G. Arvanitidis, L. K. Hansen, and S. Hauberg, "Latent Space Oddity: on the Curvature of Deep Generative Models," arXiv preprint arXiv:1710.11379 (2017).
[25] Nutan Chen, Alexej Klushyn, Richard Kurle, Xueyan Jiang, Justin Bayer, and Patrick van der Smagt, "Metrics for Deep Generative Models," arXiv preprint arXiv:1711.01204 (2017).
[26] T. Yang, G. Arvanitidis, D. Fu, X. Li, and S. Hauberg, "Geodesic Clustering in Deep Generative Models," arXiv preprint arXiv:1809.04747 (2018).
[27] J. Donahue, P. Krähenbühl, and T. Darrell, "Adversarial Feature Learning," arXiv preprint arXiv:1605.09782 (2016).
[28] Vincent Dumoulin, Ishmael Belghazi, Ben Poole, Alex Lamb, Martin Arjovsky, Olivier Mastropietro, and Aaron Courville, "Adversarially Learned Inference," arXiv:1606.00704 (2016).
[29] Zachary C Lipton and Subarna Tripathi, "Precise Recovery of Latent Vectors from Generative Adversarial Networks," arXiv preprint arXiv:1702.04782 (2017).
[30] Antonia Creswell and Anil Anthony Bharath, "Inverting the Generator of a Generative Adversarial Network," arXiv preprint arXiv:1611.05644 (2016).